\documentclass[runningheads]{llncs}

\usepackage{eccv}

\usepackage{eccvabbrv}

\usepackage{graphicx}
\usepackage{booktabs}

\usepackage{amsmath,amssymb,amsfonts}
\usepackage{array}
\usepackage{adjustbox}
\usepackage{subcaption}
\usepackage{algorithm}
\usepackage{algpseudocode}
\usepackage{multirow}
\usepackage[colorlinks,linkcolor=blue]{hyperref}

\usepackage[accsupp]{axessibility}  

\usepackage{orcidlink}
\usepackage{marvosym}

\newcommand{\customfootnotetext}[2]{{
  \renewcommand{\thefootnote}{#1}
  \footnotetext[0]{#2}}}

\begin{document}

\title{MicroAUNet: Boundary-Enhanced Multi-scale Fusion with Knowledge Distillation for Colonoscopy Polyp Image Segmentation} 

\titlerunning{Boundary-Enhanced Multi-scale Fusion with Knowledge Distillation}

\author{Ziyi Wang\orcidlink{0009-0009-0605-1224} \and
Yuanmei Zhang\orcidlink{0009-0005-9857-3062} \and
Baoying Ye\orcidlink{0000-0003-0652-8810} \and
Yimei Jiang\orcidlink{0000-0001-6123-3594} \and
Leilei Gu\orcidlink{0000-0002-3878-7262} \and
Suncheng Xiang\textsuperscript{\Letter} \orcidlink{0000-0002-9141-6460} 
}

\authorrunning{Z.~Wang et al.}

\institute{Shanghai Jiao Tong University, Shanghai 200240, China  \\
\email{chloewangzy@163.com} \\
\email{\{yuan-0102,ray01982\}@alumni.sjtu.edu.cn} \\
\email{\{eyby99,06crissi,xiangsuncheng17\}@sjtu.edu.cn}}

\maketitle

\customfootnotetext{\textsuperscript{\Letter}}{Suncheng Xiang is the corresponding author.}

\begin{abstract}
  Early and accurate segmentation of colorectal polyps is critical for reducing colorectal cancer mortality, which has been extensively explored by academia and industry. However, current deep learning-based polyp segmentation models either compromise clinical decision-making by providing ambiguous polyp margins in segmentation outputs or rely on heavy architectures with high computational complexity, resulting in insufficient inference speeds for real-time colorectal endoscopic applications. To address this problem, we propose MicroAUNet, a lightweight attention-based segmentation network, which synergistically combines depthwise-separable dilated convolutions with a single-path, parameter-shared channel-spatial attention block to effectively strengthen multi-scale boundary features. On the basis of it, a progressive two-stage knowledge-distillation scheme is introduced to transfer semantic and boundary cues from a high-capacity teacher. Extensive experiments on benchmarks also demonstrate the state-of-the-art accuracy under extremely low model complexity, indicating that MicroAUNet is suitable for real-time clinical polyp segmentation. The code is publicly available at \href{https://github.com/JeremyXSC/MicroAUNet}{https://github.com/JeremyXSC/MicroAUNet}.
\end{abstract}

\section{Introduction}

Colorectal cancer (CRC) remains one of the most prevalent and deadly malignancies worldwide~\cite{1mei2025survey,2qayoom2025polyp}. Early detection and removal of precancerous polyps during colonoscopy have proven to be the most effective strategy for reducing CRC incidence and mortality. However, traditional colonoscopy heavily relies on physician expertise, facing challenges such as a high miss rate (approximately 20-30\%) and significant operational burden due to visual fatigue and difficulties in detecting small or flat polyps~\cite{6akgol2025polyp,7sabri2025advanced}. Despite these advances, several challenges persist in clinical scenarios.

\begin{figure}[tb]
  \centering
  \includegraphics[height=5cm]{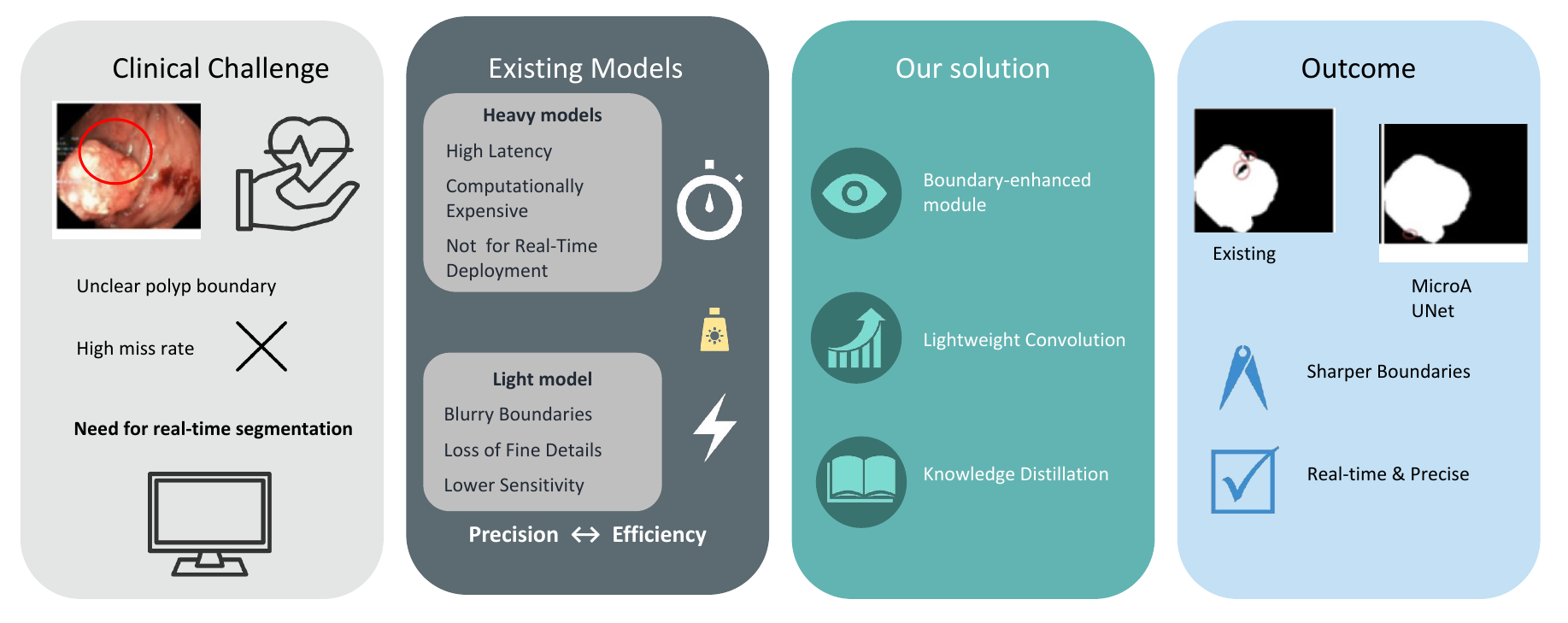}
  \caption{Motivation and overall design concept of the proposed MicroAUNet framework. It illustrates the limitations of existing polyp segmentation models—blurred boundaries and high computational costs, and this paper integrates boundary enhancement, light-weighted attention, and knowledge distillation to overcome these challenges.
  }
  \label{fig1}
\end{figure}

Deep learning-based polyp segmentation models face two key challenges: ambiguous boundary delineation due to complex backgrounds, and heavy computational costs that hinder real-time application. Consequently, developing a segmentation network that simultaneously achieves high boundary precision and computational efficiency remains an open research challenge. An overview of the motivation and design logic of our proposed framework is illustrated in Figure~\ref{fig1}.

Our proposed MicroAUNet addresses these limitations through a novel co-design that introduces a boundary-aware module to capture fine-grained details under complex backgrounds while employing a highly efficient single-path architecture with progressive distillation. This approach enables precise real-time segmentation that was previously unattainable by light-weighted models.

The main contributions of this paper can be summarised as follows.
 \begin{itemize}
\item We propose a light-weighted attention-based segmentation network named MicroAUNet, which combines depthwise separable dilated convolutions with a shared channel–spatial attention mechanism to improve boundary extraction and adaptability to polyps of diverse morphologies.
\item A progressive two-stage knowledge distillation mechanism is introduced to transfer semantic and boundary knowledge from a teacher model, enabling robust light-weighted learning with minimal computational cost.
\item Comprehensive validation on public datasets demonstrates the superiority of the proposed model in both segmentation accuracy and inference efficiency compared to existing state-of-the-art methods, highlighting its potential for clinical application. 
\end{itemize}

The remainder of this paper is structured as follows. Section~\ref{sec2} reviews related works on CNN-based, light-weighted, and attention-driven segmentation networks. In Section~\ref{sec3}, the details of the proposed MicroAUNet with reshaped blocks and a light-weighted structure are presented. Extensive evaluations compared with state-of-the-art methods and comprehensive analyses of the proposed approach are reported in Section~\ref{sec4}. Finally, Section~\ref{sec5} presents the conclusion and discussion of future work.

\section{Related Work}
\label{sec2}
\subsection{CNN-based Methods} 
Early CNN-based methods, such as FCN~\cite{12long2015fully} (pioneering end-to-end pixel-wise prediction) and U-Net~\cite{13ronneberger2015u} (introducing symmetric encoder-decoder with skip connections), established foundational architectures for medical image segmentation. Subsequent works further improved multi-scale representation: UNet++~\cite{14zhou1807nested} employed dense skip connections to mitigate semantic gaps, while ResUNet++~\cite{15jha2021comprehensive} incorporated residual and squeeze-excitation modules to enhance feature responsiveness.

Recently, boundary-aware learning has been widely adopted to refine polyp edges. For example, BMANet~\cite{wu2025bmanet} integrates boundary-guided multi-level attention, BUNet~\cite{yue2024boundary} models boundary uncertainty in ambiguous regions, PraNet~\cite{17fan2020pranet} uses reverse attention to erase non-polyp areas, and SANet~\cite{33hu2025pranet} leverages spatial attention to improve edge localisation~\cite{45fu2019dual}. Nevertheless, CNN-based methods remain limited by local receptive fields, motivating the exploration of attention and Transformer-based architectures.

\subsection{Light-weighted Segmentation Methods} 
To enable real-time colonoscopy polyp analysis, light-weighted network architectures have emerged as a key research focus. For example, UNeXt ~\cite{24valanarasu2022unext} adopted a hybrid convolution-MLP architecture, reducing the parameter count by 72$\times$ while increasing inference speed by 10$\times$, thereby providing an efficient solution for real-time diagnostic scenarios. Similarly, MALUNet~\cite{30ruan2022malunet} integrated multi-attention light-weighted modules, reducing parameters by 44$\times$ compared to U-Net while maintaining an optimal balance between computational complexity and segmentation performance in skin lesion segmentation. These light-weighted designs typically employ techniques such as depthwise separable convolutions~\cite{41chollet2017xception}, model pruning, and knowledge distillation to significantly reduce computational costs while preserving accuracy. Nevertheless, most existing light-weighted models struggle to preserve fine-grained boundary information, which is crucial for polyp segmentation tasks such as polyp detection. These limitations highlight the need for more effective feature enhancement and boundary-aware designs within compact architectures.

\subsection{Attention-based Methods} 
Recently, attention and Transformer-based architectures have gained traction in polyp segmentation due to their ability to capture global contextual dependencies. Beyond global context modelling, recent studies have also focused on progressive feature enhancement to improve multi-scale representation learning. For example, PFENet~\cite{yue2024progressive} introduces cross-stage feature enhancement modules to facilitate information interaction across different levels and progressively refine segmentation results in a coarse-to-fine manner.

However, a persistent trade-off remains between model accuracy and computational efficiency: while Transformer-based models achieve superior accuracy, their heavy architectures hinder real-time deployment; conversely, light-weighted CNNs lack boundary precision due to limited representational capacity. Motivated by this challenge, our proposed MicroAUNet aims to achieve a new balance between efficiency and accuracy through boundary-enhanced multi-scale fusion, light-weighted convolutional design, and progressive knowledge distillation.

\subsection{Knowledge Distillation}

Knowledge distillation \cite{28hinton2015distilling} addresses the growing computational cost of deep neural networks by transferring knowledge from a high-capacity \textbf{teacher model} to a compact \textbf{student model}. The core idea is to train the student to mimic the teacher's output distribution, often softened by a temperature parameter. In this work, following the standard formulation, the student network \(S\) is trained to match the teacher network \(T\) using the Kullback--Leibler (KL) divergence. The loss enables the student to capture not only the ground-truth labels but also the inter-class relationships encoded in the teacher's soft predictions, thereby achieving model compression with minimal accuracy loss. As shown in Section~\ref{sec3.3}, our proposed MicroAUNet adopts a progressive two-stage distillation strategy that extends this basic framework to incorporate both feature imitation and preference alignment.

\section{Our Method}
\label{sec3}
\subsection{Preliminary}
Colonoscopy polyp segmentation networks are commonly built on an encoder-decoder architecture. Our MicroAUNet builds upon MALUNet by adopting depthwise separable dilated convolutions and a shared attention design to enhance boundary perception and computational efficiency.

\subsection{Boundary-Enhanced Multi-Scale Feature Fusion} 

\begin{figure}[!t] 
\centering
\includegraphics[width=0.85\columnwidth]{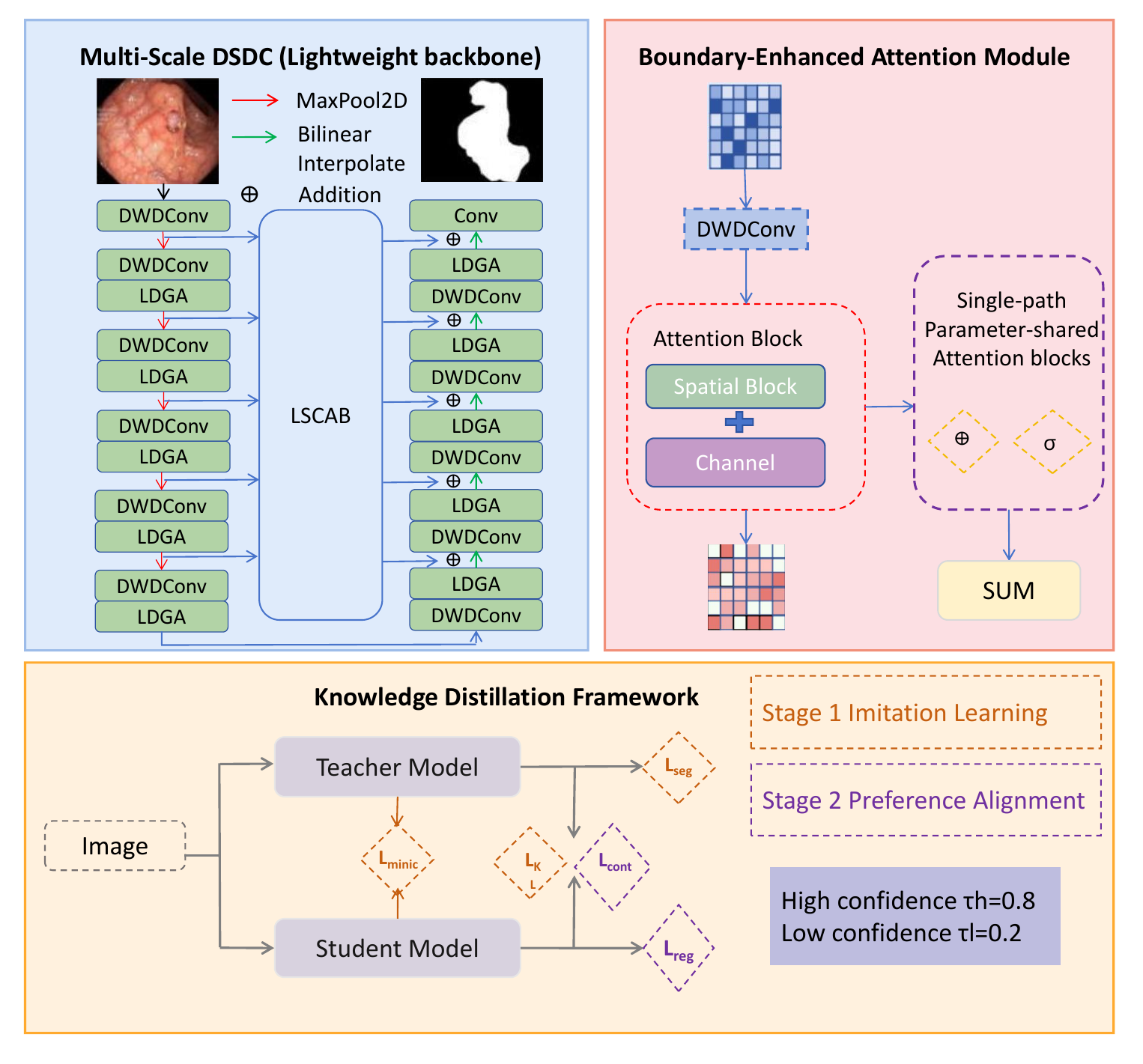}
\caption{Overview of the proposed MicroAUNet, which enhances boundary perception and achieves light-weighted segmentation through depthwise separable convolutions, shared attention, and progressive two-stage knowledge distillation.}
\label{fig2}
\end{figure}

To address the challenges of ambiguous polyp boundaries and morphological diversity, MicroAUNet incorporates a boundary-enhanced multi-scale feature fusion module. Our design builds upon MALUNet but introduces depthwise separable dilated convolutions and a shared attention design to enhance boundary perception and efficiency. An overview of the proposed architecture is demonstrated in Figure~\ref{fig2}.

\textbf{Depthwise Separable Dilated Convolutions.}
Depthwise convolution applies filters independently to each channel, while depthwise separable convolution combines depthwise and pointwise (1×1) operations to decouple spatial and channel processing, thereby reducing parameters and computation with minimal accuracy loss. 
To jointly leverage these advantages, MicroAUNet employs a depthwise separable dilated convolutional block, which decomposes convolution into two stages: dilated convolution for spatial sampling and pointwise convolution for channel fusion. This structure reduces parameter count from K$\times$K$\times$C$\times$C to K$\times$K$\times$C, achieving efficient multi-scale feature extraction crucial for delineating small or morphologically diverse polyps. \\
\textbf{Single-path and Parameter-shared Attention Block.}
Attention mechanisms are widely used to enhance feature discrimination by highlighting informative channels and salient spatial regions, enabling more accurate boundary localisation in complex endoscopic images. However, conventional multi-branch designs and independent parameterisation introduce excessive computational and memory overhead, limiting their applicability in light-weighted real-time systems. To overcome this limitation, MicroAUNet adopts a single-path spatial attention module and a parameter-shared channel–spatial attention bridge for efficient cross-scale feature integration. Specifically, the network includes two refinements: a single-path spatial attention mechanism for efficient saliency enhancement and a parameter-shared channel–spatial attention bridge for feature integration across decoder stages.

For spatial attention, a single-path depthwise separable convolution is employed to generate an attention mask. Let \(X \in \mathbb{R}^{C \times H \times W}\) denotes the input feature map. The operation \(\mathrm{DWConv}\) represents the depthwise separable dilated convolution (denoted as ``DWDConv'' in Figure~\ref{fig2}), which applies depthwise convolution with a dilation rate followed by pointwise convolution. \(\mathbf{W}_{1\times 1}\) is a pointwise convolution kernel, \(\mathrm{GELU}\) is the Gaussian Error Linear Unit activation, and \(\mathrm{Sigmoid}\) is the logistic function. The attention mask \(A\) is then computed as:
\begin{equation}
    A = \text{Sigmoid}(\mathbf{W}_{1 \times 1} \cdot \text{GELU}(\text{DWConv}(X)))
    \label{eq:placeholder_label}
\end{equation}

The attention weights are then applied to enhance salient features while retaining residual connections, where \(\odot\) denotes element-wise multiplication, and the residual addition \(X\) preserves the original feature information to avoid gradient vanishing.:
\begin{equation}
    \mathbf{X}_{\text{out}} = \mathbf{A} \odot \mathbf{X} + \mathbf{X} \label{eq:3_2}
\end{equation}

For cross-stage feature fusion, a parameter-shared spatial–channel attention bridge is introduced, which reuses shared fully connected layers across all decoder stages to minimise redundant parameters. The outputs of channel and spatial attention are then combined in a parallel weighted manner:
\begin{equation}
    \mathbf{T}^{\text{out}} = \left( \alpha \cdot \mathbf{A}_c + \beta \cdot \mathbf{A}_s \right) \odot \mathbf{T} + \mathbf{T}
    \label{eq:placeholder}
\end{equation}
where \(\mathbf{A}_c\) and \(\mathbf{A}_s\) represent channel and spatial attention, respectively, and \(\alpha ,\beta\) are learnable weighting coefficients (see Section~\ref{sec4.2} for hyperparameter settings). This parameter-sharing strategy promotes inter-stage feature consistency, enabling the model to simultaneously capture global semantics and fine-grained boundary details in complex endoscopic scenes.

\subsection{Progressive Two-Stage Knowledge Distillation}
\label{sec3.3}

Light-weighted networks typically suffer from degraded representation capacity due to limited parameters, particularly in modelling fine-grained semantics and boundaries. To address this challenge, a two-stage progressive knowledge distillation framework is introduced, where a high-capacity MALUNet serves as the teacher and MicroAUNet as the student. This framework enables the student to progressively absorb both semantic and boundary cues from the teacher during training. 

Unlike classical single-stage distillation, which simultaneously optimizes feature alignment and semantic matching, our two-stage design explicitly decouples these objectives into sequential phases. This decoupling addresses a key limitation of conventional distillation: the gradient conflict between low-level feature imitation and high-level semantic distribution matching. \\
\textbf{Stage 1: Imitation Learning.} \\
In the first stage (imitation learning), the Feature of the Student Network ($F_S$) aligns with the Feature of the Teacher Network ($F_T$) both at the feature and output distribution levels, where $F_T^l$ and $F_S^l$ denote the feature representations extracted from the $l$-th layer of the teacher and student networks, respectively, and \(\lambda_{l}\) is a layer-wise weighting coefficient. The index \(l\) ranges from 1 to 5, corresponding to the five encoder stages in the U-shaped architecture. Feature-level alignment is enforced using an $\mathcal{L}_{2}$ mimicry loss:
\begin{equation}
\mathcal{L}_{\text{mimic}} = \sum_{l=1}^{5} \lambda_{l} \left\| F_{S}^{l} - F_{T}^{l} \right\|_{2}^{2}
\end{equation}
while semantic consistency is maintained via KL divergence:
\begin{equation}
\mathcal{L}_{\text{KL}} = \frac{1}{N} \sum_{i=1}^{N} D_{\text{KL}}\!\left( T(x_{i}) \,\|\, S(x_{i}) \right)
\end{equation}
where \(T\) and \(S\) denote the teacher and student networks, respectively, \(x_i\) is the \(i\)-th input image, and \(N\) is the batch size. \(D_{\mathrm{KL}}\) represents the Kullback–Leibler divergence.
\renewcommand{\algorithmicrequire}{\textbf{Input:}}  
\renewcommand{\algorithmicensure}{\textbf{Output:}}  
\begin{algorithm}[htbp] 
\caption{Training procedure of MicroAUNet.}
\label{alg:MicroAUNet}
\begin{algorithmic}[1]
  \Require 
    Endoscopic image $I$; 
    Teacher model $\theta_T$; 
    Epochs $N$;
  \Ensure
    Optimized light-weighted model $\theta_S^{opt}$;
    
  \State $F_0 \leftarrow$ Depthwise Separable Dilated Conv($I$);
  \State $F_1 \leftarrow$ LDGA($F_0$); \quad $F_2 \leftarrow$ LSCAB($F_1$);
  \State $F_e \leftarrow$ EncoderDecoderFusion($F_2$);
  
  \For{$t = 1$ to $N$}
    \Comment{Imitation Learning} 
    \State $\mathcal{L}_{mimic} = \sum_l \lambda_l * \|F_S^l - F_T^l\|_2^2$;
    \State $\mathcal{L}_{KL} = \frac{1}{N}\sum_i D_{KL} * (T(x_i)\|S(x_i))$;
    \State $\mathcal{L}_1 = \mathcal{L}_{seg} + (1-\omega_{KL}) * \mathcal{L}_{mimic} + \omega_{KL} * \mathcal{L}_{KL}$;
    \Comment{Preference Alignment} 
    \State $\mathcal{L}_2 = L_{seg} + \mathcal{L}_{cont} + \rho \mathcal{L}_{reg}$;
    \State $\theta_S \leftarrow \text{EMA}(\theta_S, \mathcal{L}_2)$;
  \EndFor
  \State \Return $\theta_S^{opt}$ \Comment{Final optimized MicroAUNet model}
\end{algorithmic}
\end{algorithm}

The overall loss integrates segmentation loss with distillation terms:
\begin{equation}
\mathcal{L}_{1} = \mathcal{L}_{\text{seg}} + \left( 1 - \omega_{\text{KL}} \right) \cdot \mathcal{L}_{\text{mimic}} + \omega_{\text{KL}} \cdot \mathcal{L}_{\text{KL}}
\end{equation}
Specifically, the feature mimicry loss $\mathcal{L}_{\mathrm{mimic}}$ aligns intermediate feature representations to preserve boundary-related low-level cues, which is critical for accurate edge delineation. Meanwhile, the KL divergence loss $\mathcal{L}_{\mathrm{KL}}$ transfers the teacher's uncertainty knowledge at the pixel-wise prediction level, capturing inter-class relationships and ambiguous regions. These two losses are complementary and jointly enable the student to inherit both feature extraction capabilities and decision-making knowledge---a strategy widely validated in distillation literature. To balance their contributions, we adopt a curriculum learning schedule for $\omega_{\mathrm{KL}}$.

\begin{equation}
\omega_{\mathrm{KL}}(t) = \frac{1}{2} \left[ 1 + \cos\left(\pi \cdot \frac{\max(0, T-t)}{T}\right) \right]
\end{equation}

For $\omega_{\mathrm{KL}}$, we adopt a curriculum learning strategy with a cosine scheduling function that gradually shifts the training focus from feature-level alignment to semantic distribution matching. 
where $T$ is set to half of the total epochs. This allows the model to prioritize feature alignment in early epochs and KL distillation later. We will include the detailed definition and the specific value of $T$ in the revised manuscript.

\textbf{Stage 2: Preference Alignment.} 

To enable the student network to inherit the teacher's decision boundary preferences, we construct positive and negative sample sets based on the teacher's prediction confidence. Specifically, for each input image $\mathbf{x}$, we define the positive set $\mathcal{P}_{\mathbf{x}} = \{ p \mid T(\mathbf{x})_p \geq \tau_h \}$ containing high-confidence regions, and the negative set $\mathcal{N}_{\mathbf{x}} = \{ n \mid T(\mathbf{x})_n \leq \tau_l \}$ containing low-confidence regions, where $\tau_h = 0.8$ and $\tau_l = 0.2$, these thresholds were selected from common practices in knowledge distillation and contrastive learning literature.
We then randomly sample pairs $(a_p, a'_p)$ from $\mathcal{P}_{\mathbf{x}}$ as positive pairs, and $(a_p, a_n)$ from $\mathcal{P}_{\mathbf{x}} \times \mathcal{N}_{\mathbf{x}}$ as negative pairs. The contrastive loss is formulated as:

\begin{equation}
\mathcal{L}_{\mathrm{cont}} = \frac{1}{|\mathcal{A}_{\mathbf{x}}|} \sum_{(a_p, a'_p, a_n)} \left[ \left\| S(\mathbf{x})_{a_p} - S(\mathbf{x})_{a'_p} \right\|_2^2 - \left\| S(\mathbf{x})_{a_p} - S(\mathbf{x})_{a_n} \right\|_2^2 + m \right]_+
\end{equation}

While the imitation learning stage enables the student to align its feature representations with the teacher, it does not explicitly enforce the student to inherit the teacher's decision boundaries, particularly in ambiguous regions where polyp boundaries are difficult to distinguish from background. To address this, we introduce a contrastive loss that encourages the student to learn from the teacher's confidence-based preferences. Specifically, the contrastive loss enhances intra-class compactness by pulling together features of high-confidence positive samples, while improving inter-class separability by pushing apart features of positive and negative samples. This helps the student refine its decision boundaries in complex scenarios and better capture the teacher's discriminative knowledge.

Given the distribution characteristics of polyp datasets, our optimization objective is to pull similar samples closer (intra-class compactness) and push dissimilar samples apart (inter-class separation) in the feature space \cite{chopra2005learning}. 
The final objective is defined as:
\begin{equation}
    \mathcal{L}_2 = \mathcal{L}_{\text{seg}} + \mathcal{L}_{\text{cont}} + \rho \cdot \mathcal{L}_{\text{reg}}
    \label{eq:placeholder7}
\end{equation}
where \(\mathcal{L}_{\mathrm{cont}}\) encourages consistency among positive samples and differentiation from negatives, and \(\mathcal{L}_{\mathrm{reg}}\) acts as a regularisation term to prevent overfitting, with \(\rho\) as the regularization coefficient (see Section~4.2 for hyperparameter settings). This two-stage progressive distillation framework enables MicroAUNet to achieve strong boundary delineation and semantic understanding despite its compact parameter space.
A complete training procedure of MicroAUNet is illustrated in Algorithm~\ref{alg:MicroAUNet}.

\section{Experiments}
\label{sec4}

\subsection{Datasets and Evaluation Metrics}
\label{sec4.1}

To comprehensively evaluate the proposed MicroAUNet model, we conducted extensive experiments on two public polyp segmentation datasets: Kvasir-SEG~\cite{31jha2019kvasir} and CVC-ClinicDB~\cite{32bernal2015wm}. All models were trained under the same hyperparameter settings outlined in Section~\ref{sec4.2} to ensure a fair comparison. The evaluation encompasses both comparative analysis against state-of-the-art methods and an internal ablation study to assess the individual contribution of each proposed component. Performance was evaluated using five standard metrics: mean Dice coefficient (mDice), mean Intersection over Union (mIoU), Accuracy (Acc), Specificity (Spe), and Sensitivity (Sen).

\subsection{Implementation details}
\label{sec4.2}
During the experiment, training was conducted for 300 epochs with a batch size of 8. The AdamW optimiser was employed with an initial learning rate of 0.001, and a cosine annealing learning rate scheduler was applied to improve training stability. To ensure reproducibility, the random seed was fixed at 42. All experiments were conducted on a platform equipped with an Intel(R) Xeon(R) Gold 6330 CPU @ 2.1 GHz with 128 GB of RAM and an NVIDIA GeForce RTX 3090 GPU with 24 GB of memory. 

The hyperparameters used in our proposed method are as follows. For the parallel attention fusion in Eq.~\ref{eq:placeholder}, the learnable weights \(\alpha\) and \(\beta\) are both initialized to \(0.5\). For the two-stage knowledge distillation, the high-confidence threshold \(\tau_{h}\) and low-confidence threshold \(\tau_{l}\) are set to \(0.8\) and \(0.2\), respectively. The regularization coefficient \(\rho\) in Eq.~\ref{eq:placeholder7} is set to \(0.3\).

\subsection{Comparison with State-of-the-Arts}

\begin{figure}[!t] 
\centering
\includegraphics[width=0.90\columnwidth]{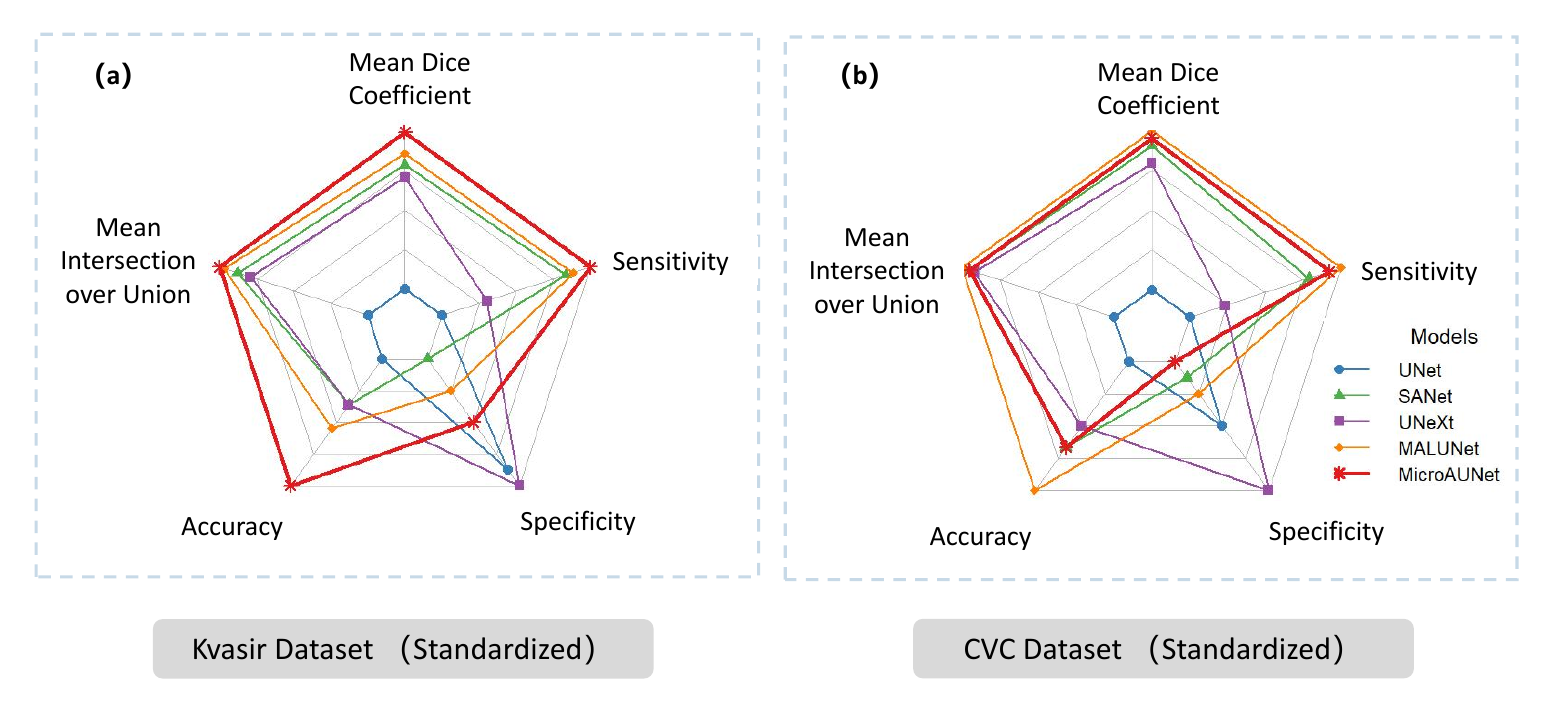}
\caption{Quantitative comparison of different segmentation models on Kvasir and CVC datasets. Radar charts visualise the performance (mDice, mIoU, Accuracy, Specificity, and Sensitivity) of UNet, SANet, UNeXt, MALUNet, and the proposed MicroAUNet, demonstrating the superior accuracy–efficiency trade-off of MicroAUNet.}
\label{fig3}
\end{figure}

To comprehensively evaluate the proposed MicroAUNet, experiments were conducted on the Kvasir-SEG and CVC-ClinicDB datasets, with comparisons against several representative segmentation models, including UNet, SANet, UNeXt, and MALUNet. The results are presented in Table~\ref{table1} and visualised by radar charts in Figure~\ref{fig3}.
\begin{table}[htbp] 
  \centering
  \caption{Performance comparison of different models on Kvasir and CVC datasets, normalised scores are shown for Kvasir Dataset and CVC Dataset.}
  \small
  \setlength{\tabcolsep}{0.6mm}{
    \begin{tabular}{lcccccccccc}
    \toprule
    \multirow{2}{*}{Model} & \multicolumn{5}{c}{Kvasir} & \multicolumn{5}{c}{CVC} \\
    \cmidrule(lr){2-6} \cmidrule(lr){7-11}
    & mDice & mIoU & Acc & Spe & Sen & mDice & mIoU & Acc & Spe & Sen \\
    \midrule
    UNet & 0.818 & 0.746 & 0.940 & 0.967 & 0.859 & 0.823 & 0.750 & 0.953 & 0.966 & 0.860 \\
    SANet & 0.886 & 0.847 & 0.944 & 0.960 & 0.895 & 0.898 & 0.869 & 0.957 & 0.963 & 0.901 \\
    UNeXt & 0.879 & 0.837 & 0.944 & \textbf{0.968} & 0.872 & 0.889 & 0.865 & 0.956 & \textbf{0.970} & 0.872 \\
    MALUNet & 0.892 & 0.857 & 0.946 & 0.962 & 0.897 & \textbf{0.906} & \textbf{0.875} & \textbf{0.959} & 0.964 & \textbf{0.912} \\
    MicroAUNet & \textbf{0.904} & \textbf{0.861} & \textbf{0.951} & 0.964 & \textbf{0.902} & 0.902 & 0.869 & 0.957 & 0.962 & 0.908 \\
    \bottomrule
    \end{tabular}}%
  \label{table1}%
\end{table}%

The results demonstrate that MicroAUNet achieves the best trade-off between segmentation accuracy and efficiency, and generally outperforms other light-weighted and full-capacity models across both datasets. The baseline UNet performs significantly worse than the advanced architectures, achieving only 0.818/0.746 (mDice/mIoU) on Kvasir and 0.823/0.750 on CVC. While SANet and UNeXt exhibit moderate improvements through attention and MLP hybridisation, MicroAUNet attains the best trade-off between accuracy and efficiency. Specifically, MicroAUNet achieves 0.904/0.861 (mDice/mIoU) on Kvasir and 0.902/0.869 on CVC, surpassing MALUNet with substantially fewer parameters (0.0249M vs. 0.175M). The relative performance of MicroAUNet and MALUNet varies across the two datasets, which may be attributed to differences in data distributions and architecture-specific feature representations. Although MicroAUNet is slightly inferior to MALUNet on CVC-ClinicDB, it maintains competitive segmentation accuracy with only 0.0249M parameters, compared with 0.175M for MALUNet, demonstrating a favorable accuracy–compactness trade-off.

\begin{figure}[!t] 
\centering
\includegraphics[width=0.90\columnwidth]{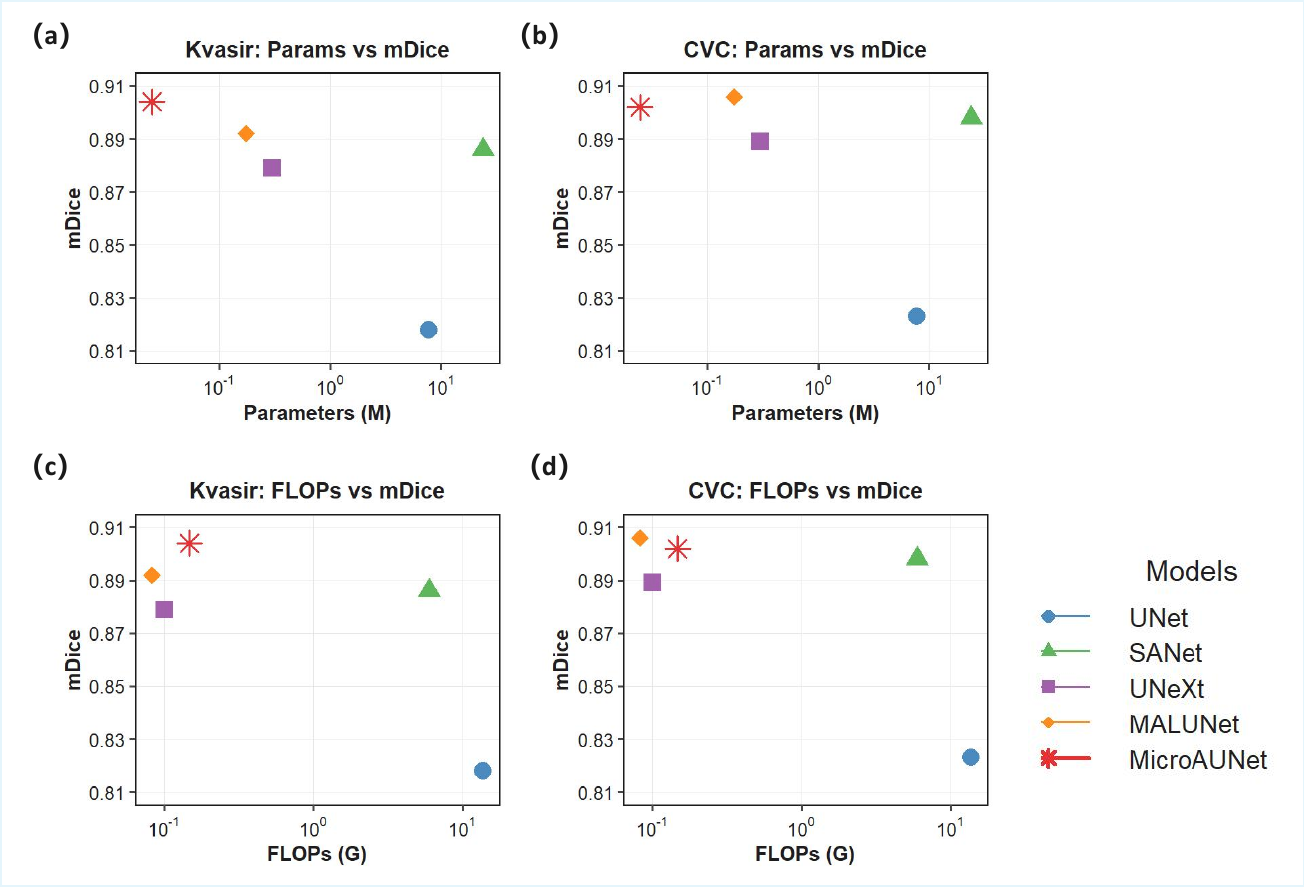}
\caption{Comprehensive analysis of efficiency and performance across models. Scatter plots compare parameter count and computational complexity (Params and FLOPs) versus segmentation accuracy, illustrating that MicroAUNet achieves the optimal balance among light-weighted and full-scale architectures.}
\label{fig5}
\end{figure}

The model complexity analysis is presented in Table~\ref{table2}. MicroAUNet has only 0.0249M parameters, which is substantially fewer than UNet (7.77M) and SANet (23.90M). Although its GFLOPs (0.148) are slightly higher than MALUNet (0.083), MicroAUNet maintains extremely low model complexity while achieving competitive segmentation accuracy, demonstrating a favorable accuracy–parameter trade-off. Figure~\ref{fig5} further supports these findings. These findings confirm the effectiveness of the proposed light-weighted architectural design, which trades off computational redundancy for improved efficiency while retaining robust representational capacity.

\begin{table}[!t] 
  \centering
  \caption{Comparison of Model complexity in terms of Params(M) and FLOPs (G).}
  \small
  \setlength{\tabcolsep}{2mm}{
    \begin{tabular}{lccccc}
    \toprule
    Model & UNet & SANet & UNeXt & MALUNet & \textbf{MicroAUNet} \\
    \midrule
    Params (M) & 7.77 & 23.90 & 0.30 & 0.175 & \textbf{0.0249} \\
    FLOPs (G)  & 13.78 & 5.99 & 0.10 & 0.083 & \textbf{0.148} \\
    \bottomrule
    \end{tabular}}%
  \label{table2}%
\end{table}%

Removing the light-weighted strategy in MicroAUNet-1 led to a 1.34\% decrease in mDice (0.904 → 0.892) and a 0.47\% drop in mIoU (0.861 → 0.857) on the Kvasir dataset, demonstrating that the light-weighted module effectively improves segmentation accuracy through parameter compression and feature reuse. However, on the CVC dataset, MicroAUNet-1 slightly outperformed the full model in mDice (0.906 vs. 0.902). This may be attributed to the fact that lesions in the CVC dataset have clearer boundaries, where the marginal feature loss from light-weighted convolution has a limited impact on segmentation performance.

In contrast, MicroAUNet-2, which excludes the imitation learning stage, exhibited a notable decline in performance. The mDice dropped by 3.21\% (0.904 → 0.875) on Kvasir, and the mIoU decreased by 2.71\% (0.869 → 0.846) on CVC. These results confirm that imitation learning via teacher–student knowledge distillation substantially enhances the model’s ability to distinguish subtle lesions in medical images.

When removing the preference contrastive learning mechanism (MicroAUNet-3), the mDice and mIoU decreased by 3.43\% and 2.91\%, respectively, on the Kvasir dataset. This indicates that the contrastive learning strategy strengthens intra-class feature consistency and improves robustness in complex backgrounds.

Overall, the proposed MicroAUNet achieves superior performance over all variants ablated across both data sets, reflecting the complementary and synergistic effects between the light-weighted design and the training strategy. The light-weighted modules reduce redundant parameters, while imitation and contrastive learning compensate for representational loss through feature-guided supervision.

Consequently, the complete model achieves a maximum mDice of 0.904 on Kvasir and an mIoU of 0.869 on CVC, outperforming MicroAUNet-3 by 3.09\%. Moreover, MicroAUNet exhibits the highest cross-dataset stability, with an mDice standard deviation of 0.0014 (0.904 vs. 0.902), which is notably lower than that of MicroAUNet-2 (0.0015) and MicroAUNet-3 (0.0040). These findings demonstrate that multi-component collaboration effectively enhances segmentation stability and robustness across datasets.

\subsection{Ablation studies}
To investigate the contributions of individual components, ablation studies were performed on the Kvasir and CVC-ClinicDB datasets, the results are shown in Table~\ref{table:microaunet_comparison}. Three variants were designed for comparison: MicroAUNet-1 (without depthwise separable dilated convolutions), MicroAUNet-2 (without the imitation learning stage), and MicroAUNet-3 (without the preference alignment stage).

\begin{table}[!t] 
  \centering
  \caption{Ablation Study of proposed MicroAUNet method on the Kvasir and CVC-ClinicDB datasets.}
  \small
  \setlength{\tabcolsep}{0.6mm}{
    \begin{tabular}{lcccccccccc}
    \toprule
    \multirow{2}{*}{Model} & \multicolumn{5}{c}{Kvasir} & \multicolumn{5}{c}{CVC} \\
    \cmidrule(lr){2-6} \cmidrule(lr){7-11}
    & mDice & mIoU & Acc & Spe & Sen & mDice & mIoU & Acc & Spe & Sen \\
    \midrule
    MicroAUNet-1 & 0.892 & 0.857 & 0.946 & 0.960 & 0.896 & \textbf{0.906} & \textbf{0.875} & \textbf{0.960} & \textbf{0.966} & \textbf{0.911} \\
    MicroAUNet-2 & 0.875 & 0.840 & 0.943 & 0.957 & 0.892 & 0.878 & 0.846 & 0.951 & 0.956 & 0.903 \\
    MicroAUNet-3 & 0.873 & 0.836 & 0.941 & 0.955 & 0.891 & 0.881 & 0.843 & 0.953 & 0.959 & 0.906 \\
    MicroAUNet   & \textbf{0.904} & \textbf{0.861} & \textbf{0.951} & \textbf{0.964} & \textbf{0.902} & 0.902 & 0.869 & 0.957 & 0.962 & 0.908 \\
    \bottomrule
    \end{tabular}}%
  \label{table:microaunet_comparison}%
\end{table}%

 \subsection{Qualitative Analysis}
 
\begin{figure}[!t] 
\centering
\includegraphics[width=1.00\columnwidth]{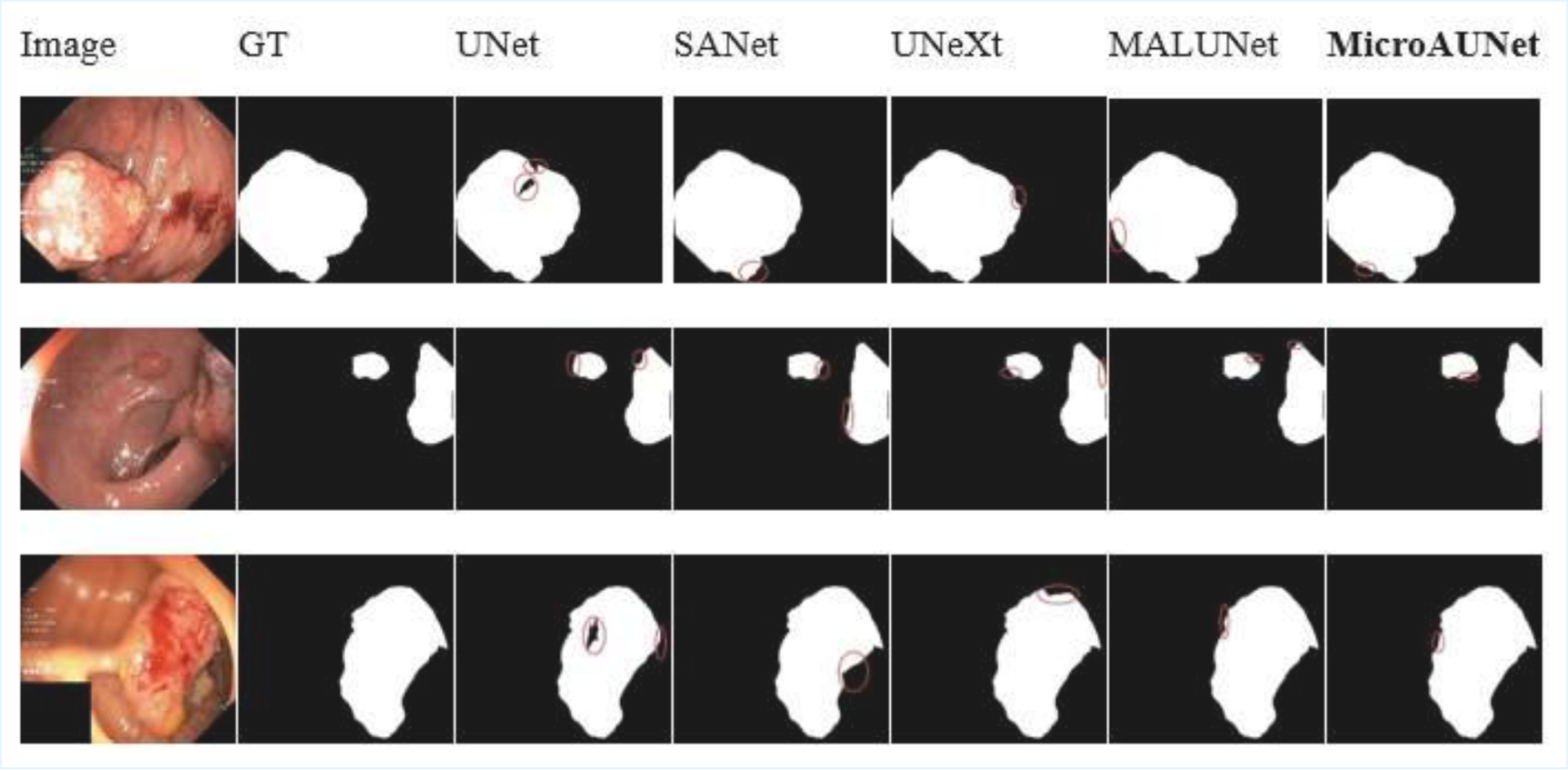}
\caption{Visual comparison between the proposed method and four state-of-the-art ones. Visual examples show that MicroAUNet produces sharper, more continuous boundaries and fewer background artefacts compared with UNet, SANet, UNeXt, and MALUNet, yielding results closest to the ground truth.}
\label{fig4}
\end{figure}

 Figure~\ref{fig4} shows qualitative comparisons among Ground Truth (GT), U-Net, SANet, UNeXt, MALUNet, and the proposed MicroAUNet. As seen in the first two models, U-Net and SANet miss part of the polyp edges or produce fragmented masks. UNeXt and MALUNet generate more complete results but still blur fine boundaries in low-contrast areas. In contrast, MicroAUNet accurately delineates polyp contours and maintains boundary continuity, even under occlusion or irregular shapes. The red circles highlight that MicroAUNet removes background artefacts and better preserves small structures, demonstrating clearer and more reliable segmentation consistent with ground truth.
 
\subsection{Discussion and Limitation}

The proposed MicroAUNet achieves an effective balance between segmentation precision and computational efficiency, making it well-suited for real-time clinical applications. In contrast, MicroAUNet employs depthwise separable dilated convolutions and a parameter-shared single-path attention design, enabling efficient multi-scale boundary extraction with minimal redundancy.

While light-weighted models typically risk losing representational capacity, the introduced progressive two-stage knowledge distillation successfully transfers both semantic and boundary knowledge from a teacher model, preserving high segmentation accuracy under a limited number of parameters. 
Experiments on Kvasir and CVC-ClinicDB demonstrate that MicroAUNet surpasses state-of-the-art models in mDice and mIoU while using fewer parameters and achieving faster inference. Nonetheless, its performance may decline on extremely low-quality or unseen imaging styles.

Although MicroAUNet achieves a favourable balance between accuracy and efficiency, some limitations remain. Its generalisation to unseen clinical environments, such as images from different endoscopic devices or lighting conditions, has not been fully validated. While strong performance is shown on Kvasir and CVC-ClinicDB, evaluation on multi-centre or real-world datasets is needed to confirm broader robustness. Additionally, the two-stage distillation framework relies on the teacher model’s quality; biased or noisy teacher representations may limit the student network’s performance. 
Beyond still-frame segmentation, MicroAUNet can be extended to continuous colonoscopy video by incorporating temporal information across adjacent frames. Temporal consistency constraints and frame-to-frame feature or mask propagation could reduce mask fluctuations and improve prediction continuity, while lightweight tracking may further handle camera motion and temporary occlusion. Future work will also evaluate per-frame latency and throughput on continuous colonoscopy streams to more directly assess real-time applicability.

\section{Conclusion}
 \label{sec5}
In this work, we proposed MicroAUNet, a boundary-enhanced lightweight segmentation model that achieves competitive segmentation accuracy under an extremely compact parameter budget. Comprehensive experiments verify that MicroAUNet achieves state-of-the-art segmentation performance with substantially reduced computational cost, demonstrating strong potential for real-time colonoscopic deployment. Future research will focus on exploring domain adaptation and contrastive learning to enhance cross-domain generalisation across diverse imaging devices and clinical conditions.

\section*{Acknowledgement}
This work was partially supported by the National Natural Science Foundation of China under Grant No.62301315 and Shanghai Municipal Health Commission Health Industry Clinical Research Special Project under Grant No.202340010.

%
%
\bibliographystyle{splncs04}
\bibliography{sn-bibliography}
\end{document}